\documentclass[english]{article}
\usepackage[T1]{fontenc}
\usepackage[latin9]{inputenc}
\usepackage{geometry}
\makeatletter
\usepackage{url}
\usepackage{lipsum}  
\usepackage{graphicx}
\usepackage{parskip}
\usepackage[style=apa, sorting=nyt]{biblatex}
\usepackage{float}
\usepackage{placeins}
\usepackage{amsmath}
\usepackage{latexsym}
\usepackage{array} 
\usepackage{wrapfig}
\usepackage{tabularx}
\usepackage{amsmath, amssymb}
\usepackage{csquotes}
\usepackage{subcaption}

\makeatother

\usepackage{babel}
\begin{document}

\title{
    \vspace*{2cm}
    \textbf{Enhancing ML Interpretability for Credit Scoring }\\
    \vspace*{1cm}
}

\author{
    \textbf{Sagi Schwartz}$^{1}$, 
    \textbf{Qinling Wang}$^{2}$, 
    \textbf{Fang Fang}$^{2,3}$\\[1ex]
    $^1$Dept. of Computer Science, Delft University of Technology\\
    $^2$Dept. of Applied Mathematics, Delft University of Technology\\
    $^3$FF Quant Advisory B.V., Utrecht\\
    \texttt{s.schvarcz@student.tudelft.nl}, \texttt{q.wang-7@tudelft.nl}, \texttt{f.fang@tudelft.nl}
}

\date{}

\maketitle
\thispagestyle{empty}

\let\clearpagebackup\clearpage
\renewcommand{\clearpage}{ }

\onecolumn

\vspace*{1.5cm}

\vspace*{2cm}

\noindent
{\small
Word count: 4813\\
Number of figures and tables: 8\\
}
\vfill


\twocolumn
\let\clearpage\clearpagebackup  
\clearpage
\setcounter{page}{1}

\onecolumn

\begin{abstract}
Predicting default is essential for banks to ensure profitability and financial stability. While modern machine learning methods often outperform traditional regression techniques, their lack of transparency limits their use in regulated environments. Explainable artificial intelligence (XAI) has emerged as a solution in domains like credit scoring. However, most XAI research focuses on post-hoc interpretation of black-box models, which does not produce models lightweight or transparent enough to meet regulatory requirements, such as those for Internal Ratings-Based (IRB) models.

This paper proposes a hybrid approach: post-hoc interpretations of black-box models guide feature selection, followed by training glass-box models that maintain both predictive power and transparency.

Using the Lending Club dataset, we demonstrate that this approach achieves performance comparable to a benchmark black-box model while using only 10 features - an 88.5\% reduction. In our example, SHapley Additive exPlanations (SHAP) is used for feature selection, eXtreme Gradient Boosting (XGBoost) serves as the benchmark and the base black-box model, and Explainable Boosting Machine (EBM) and Penalized Logistic Tree Regression (PLTR) are the investigated glass-box models.

We also show that model refinement using feature interaction analysis, correlation checks, and expert input can further enhance model interpretability and robustness.
\end{abstract}


\section*{Key Messages}

\begin{itemize} 
  \item Transparent, lightweight ML models better meet regulatory and practical needs. 
  \item Our approach can cut features by 80\%+ without reducing predictive accuracy.
  \item Post-hoc tools are used for selecting key features from black-box models.
  \item The resulting reduced glass-box models  ensure transparency and accuracy.
\end{itemize}

\section{Introduction}
Credit scoring is one of the core tasks in the financial sector. While traditional statistical models remain widely used, there is a growing trend among banks to incorporate artificial intelligence (AI) into credit scoring processes. As of November 2024, 54.12\% of European banks were leveraging AI for credit scoring and creditworthiness assessment \parencite{EBA2024}. Regarding the performance of AI based models, \textcite{alonso_machine_2020, alonso_understanding_2021} reported that machine learning (ML) techniques can significantly improve predictive power, potentially leading to a 12.4\% to 17\% reduction in regulatory capital requirements.

Despite these advances, logistic regression and other generalized linear models  remain dominant in probability of default (PD) modeling. This is primarily due to their inherent interpretability, which is a desirable and sometimes mandatory property for models used in regulated environments, as emphasized in legal frameworks such as the EU GDPR Recital 71 and the U.S. Equal Credit Opportunity Act.

Recent studies have  applied ML models extensively to credit scoring tasks \parencite{misheva_explainable_2021, ariza-garzon_explainability_2020, de_lange_explainable_2022}, often using post-hoc explanation methods such as SHapley Additive exPlanations (SHAP) \parencite{lundberg_unified_2017} and Local Interpretable Model-agnostic Explanations (LIME) \parencite{ribeiro_why_2016}. These tools help interpret model predictions by identifying important features and explaining model behavior in ways that align with domain knowledge and human intuition. For example, top-ranked features identified via SHAP often correspond to variables already recognized by practitioners as relevant financial indicators.

However, despite these interpretive efforts, the inherent complexity of black-box ML models remains a major obstacle. These models typically involve numerous feature interactions and exhibit non-linear relationships that are difficult to understand and communicate, particularly to model owners and regulatory supervisors. There are two primary sources of this complexity: the high dimensionality of input data and the structural intricacies of the algorithms themselves. This paper proposes a framework that addresses both.

Our approach begins by training a high-performing black-box ML model using the full feature set, which serves as both a performance benchmark and the basis for feature selection. We then apply a post-hoc interpretation tool, specifically SHAP in our testing example, to rank the features by importance. A subset of top-ranked features is selected and used to train interpretable, glass-box ML models, such as the Explainable Boosting Machine (EBM) and Penalized Logistic Tree Regression (PLTR).
%
Finally, the resulting simple models can be fine-tuned via feature interaction analysis, correlation analysis, and expert opinion, for further enhancement of the model performance.

This approach ultimately produces explainable and technically transparent models that align more closely with the expectations of both regulators and practitioners in the credit risk domain.

\section{Industry Benchmark: Logistic Regression}

Laying the groundwork for high-performance interpretable machine learning algorithms, we begin by examining logistic regression, the preferred modeling technique in the industry. Logistic regression is a statistical model that estimates the log-odds of an event occurring as a linear combination of independent variables. The estimated log-odds is then transformed into a probability that the event will take place. By setting a threshold for this probability, logistic regression can be used as a binary classifier, where outcomes above and below the threshold correspond to two distinct classes.

Briefly, logistic regression is a generalized linear model that estimates the probability of a binary outcome:
\[
p = P(Y=1 \mid \mathbf{X}=\mathbf{x}),
\]
where \(\mathbf{X} = (X_1, X_2, \dots, X_d)\) is the vector of features and \(\mathbf{x} = (x_1, x_2, \dots, x_d)\) is a realization of these features. The logistic regression model provides an \textit{estimate} of \(p\) given by
\begin{equation}\label{eq:logi-reg}
\hat{p} \triangleq \sigma\big(z(\boldsymbol{\beta},\mathbf{x})\big) 
= \frac{1}{1+e^{-z(\boldsymbol{\beta},\mathbf{x})}},
\end{equation}
where \(\sigma(\cdot)\) is the sigmoid function and the linear component is
\[
z(\boldsymbol{\beta},\mathbf{x}) = \beta_0 + \beta_1x_1 + \beta_2x_2 + \cdots + \beta_dx_d,
\]
with parameter vector \(\boldsymbol{\beta} = (\beta_0, \beta_1, \dots, \beta_d)\).

The model is typically trained by minimizing the negative log-likelihood (log-loss) function:
\begin{equation}\label{eq:log-loss}
\mathcal{L}(\boldsymbol{\beta}) 
= -\frac{1}{n}\sum_{i=1}^{n}\Big[ y_i \ln \hat{p}_i + (1-y_i)\ln(1-\hat{p}_i)\Big],
\end{equation}
where \(n\) is the number of training samples, \(y_i \in \{0,1\}\), \(\hat{p}_i = \sigma(z(\boldsymbol{\beta}, \mathbf{x}_i))\), and \(\mathbf{x}_i\) is the \(i\)-th feature vector. This cost function penalizes confident but incorrect predictions. Optimization is usually performed via gradient descent or related algorithms to obtain parameter estimates \(\hat{\boldsymbol{\beta}}\).

One of the key advantages of logistic regression is interpretability. By rearranging Eq.~(\ref{eq:logi-reg}) in terms of \textit{fitted log-odds}, we have
\[
\ln\!\left(\frac{\hat{p}}{1-\hat{p}}\right) = z(\boldsymbol{\beta}, \mathbf{x}) = \beta_0 + \beta_1x_1 + \dots + \beta_dx_d.
\]
This expresses a linear relationship between the features and the log-odds of the positive class. Holding all other features constant, a one-unit increase in feature \(x_i\) multiplies the odds of being in the positive class by \(e^{\beta_i}\). The intercept \(\beta_0\) represents the log-odds when all features \(x_1, x_2, \dots, x_d\) are zero, corresponding to odds of \(e^{\beta_0}\).

However, a key limitation of logistic regression is  that the resulting model is often too simple to accurately capture complex data patterns, which may lead to poor performance compared to more flexible ML models \parencite{alonso_understanding_2021, alonso_machine_2020, de_lange_explainable_2022, schmitt_explainable_2024, moscatelli_corporate_2019}.

\section{Existing Explainable AI Models}

There have been continuous advancements both in terms of complexity and capability of machine learning algorithms in recent years. On the one hand, these algorithms enable faster and more accurate data modeling; on the other, they pose challenges in terms of explainability and trust. 

In the existing literature, there are two main approaches in interpreting and explaining ML models. The first approach is to use inherently interpretable methods - also known as glass-box models - such as linear regression, decision trees, and generalized additive models (GAMs). While these methods offer desirable transparency, their predictive performance may be insufficient. The second approach involves model-agnostic interpretation tools that can be applied to any supervised machine learning model, also known as post-hoc methods.

In the subsections that follow, we briefly recall some key post-hoc explainable AI (XAI) methods, along with recent advances in glass-box models reported in the literature. These will lay the foundation for the approach we will propose in Section \ref{sec:method}.

\subsection{Post-Hoc Explanation Methods}

In the literature of credit scoring, two post-hoc explainable XAI methods are mostly studied: Locally Interpretable Model-Agnostic Explanations (LIME) and SHapley Additive exPlanations (SHAP). 

LIME, introduced by \cite{ribeiro_why_2016}, provides explanations for individual predictions by fitting a simple local model around the observation being explained, using similar data points. The local model is typically chosen from interpretable classes such as linear models or decision trees. 

SHAP, introduced by \cite{lundberg_unified_2017}, computes Shapley values and presents them as an additive feature attribution method, which is effectively a linear model. A notable advantage of SHAP is its visualization tools, including the SHAP summary plot and dependence plot, which are very handy to use and greatly enhance the interpretation of the model outcomes. 

Other interpretation methods include Partial Dependence Plots, Accumulated Local Effects (ALE) Plots, Functional Decomposition, and Permutation Feature Importance, among others. We will not explore these methods in this paper but refer the reader to \cite{molnar2022interpretable} for relevant details.

Here we elaborate further on SHAP, one of the most widely used methods for providing global explanations of black-box models and a key component of our experiments. SHAP is based on the Shapley values from game theory. More precisely, it uses Shapely values to determine the contribution of each feature to a final prediction made by a model. 

Let $F = \{1,2,\dots,d\}$ be the set of indices for all features, and let $\mathbf{x} \in \mathbb{R}^d$ be a feature vector. For a given feature with number $j \in F$, the \emph{Shapley value} of this feature is defined as
\[
\phi_j(\mathbf{x}) \;=\; 
\sum_{S \subseteq F \setminus \{j\}} 
\frac{|S|!\,(|F|-|S|-1)!}{|F|!}\;
\Big[\, f_{S \cup \{j\}}(\mathbf{x}_{S \cup \{j\}}) - f_S(\mathbf{x}_S)\,\Big],
\]
where $|S|$ is the cardinality of the subset $S$, i.e. the number of elements in $S$;  $\mathbf{x}_S \in \mathbb{R}^{|S|}$ is the subvector of $\mathbf{x}$ restricted to the indices in $S$; $f_S : \mathbb{R}^{|S|} \to \mathbb{R}$ denotes the model restricted to the feature set $S$ (with all other features marginalized out or fixed according to a baseline). The weight $
    w(S) = \frac{|S|!\,(|F|-|S|-1)!}{|F|!}$
is the fraction of feature orderings in which the subset $S$ appears before $j$.

The computational complexity for Shapley values grows exponentially in the number of features. To address this issue, SHAP provides two efficient approximation methods for Shapley values: model agnostic Kernel SHAP and tree-based Tree SHAP. The latter, introduced in \cite{lundberg_consistent_2019},   dramatically reduces computation time by exploiting the structure of tree-based models.

\subsection{Black-box Models with Post-Hoc Explanations}\label{sec:XGBoost}

One of the best performing black-box models for credit scoring - as well as for other applications in finance such as time-series prediction - is XGBoost (eXtreme Gradient Boosting). \textcite{alonso_understanding_2021} showed that XGBoost, along with random forest, consistently outperformed other models  examined in terms of AUC-ROC. In addition, \cite{jones_empirical_2015} demonstrated that gradient boosting trees were among the top-performing ML classifiers for the credit scoring task conducted in their study. Similar results were also reported by \textcite{de_lange_explainable_2022} and \textcite{moscatelli_corporate_2019}.

XGBoost is an optimized implementation of the gradient boosting framework developed by \textcite{chen_xgboost_2016}. It constructs an ensemble of $K$ regularized Classification
And Regression Trees (CARTs) to solve supervised learning tasks, including regression and classification. The prediction for an instance $\mathbf{x}_i$ is given by 
\begin{equation}\label{eq:XGBoost}
\hat{y}_i = \sum_{k=1}^{K} f_k(\mathbf{x}_i), \quad f_k \in \mathcal{F},
\end{equation}
where $K$ is the number of trees, and $\mathcal{F}$ is the space of all CARTs. Each tree $f_k$ maps the input feature vector $\mathbf{x}_i \in \mathbb{R}^d$ to a predicted score, and the ensemble prediction is obtained by summing the outputs of all trees. 
The model is trained to minimize a regularized objective function
\[
\mathcal{L}(\phi) = \sum_{i=1}^n l(y_i, \hat{y}_i) + \sum_{k=1}^K \Omega(f_k).
\]
Here $l$ is a differentiable convex loss function that quantifies the difference between the ground truth $y_i$ and the prediction $\hat{y}_i$. It is squared error for regression and logistic loss for classification. For binary classification, the logistic loss corresponds to the negative log-likelihood, which is very similar to Eq. (\ref{eq:log-loss}):
\[
l(y_i, \hat{y}_i) = - \Big[ y_i \ln \sigma(\hat{y}_i) + (1-y_i) \ln \big(1 - \sigma(\hat{y}_i)\big) \Big],
\]
where $\sigma(\cdot)$  is again the sigmoid function. 

The second term, $\Omega(f)$, penalizes model complexity such that the model favors simpler but predictive functions, and is given by
$
\Omega(f) = \gamma T + \frac{1}{2} \lambda \|w\|^{2},
$
where $T$ is the number of leaves in the tree, $w$ is the vector of leaf weights, and  $\gamma$ and $\lambda$ are respective regularization parameters. 

The objective function above is equivalent to
$
\mathcal{L}^{(t)} = \sum_{i} l(y_i, \hat{y}^{(t-1)}_i + f_t(\mathbf{x}_i)) + \Omega(f_t),
$
with $f_t(\mathbf{x}_i)$ being the CART added at iteration $t$. To simplify the optimization, a second-order Taylor expansion is applied around the current prediction $\hat{y}^{(t-1)}_i$, which gives
\[
\mathcal{L}^{(t)} \simeq \sum_{i} \Big[ l(y_i, \hat{y}^{(t-1)}_i) + g_i f_t(\mathbf{x}_i) + \tfrac{1}{2} h_i f_t^2(\mathbf{x}_i) \Big] + \Omega(f_t),
\]
where 
\[
g_i = \frac{\partial l(y_i, \hat{y}^{(t-1)}_i)}{\partial \hat{y}^{(t-1)}_i}, 
\quad 
h_i = \frac{\partial^2 l(y_i, \hat{y}^{(t-1)}_i)}{\partial (\hat{y}^{(t-1)}_i)^2}.
\]
Dropping constant terms independent of $f_t$ yields the simplified objective
\[
\widetilde{\mathcal{L}}^{(t)} \simeq \sum_{i} \Big[ g_i f_t(\mathbf{x}_i) + \tfrac{1}{2} h_i f_t^2(\mathbf{x}_i) \Big] + \Omega(f_t).
\]
For the logistic loss, these derivatives are explicitly
\[
g_i = \sigma(\hat{y}^{(t-1)}_i) - y_i, 
\quad 
h_i = \sigma(\hat{y}^{(t-1)}_i) \big(1 - \sigma(\hat{y}^{(t-1)}_i)\big).
\]

Since gradient boosting models consist of an ensemble of decision trees, where each tree may utilize multiple features to make predictions, it is challenging to gauge the contribution of individual features to the model's predictions.

To address this model transparency issue, XGBoost library is equipped with several feature importance metrics that can be used to obtain more knowledge on the model's inner workings.
The most commonly used metric is the {\em gain}, which measures how much a feature improves the model's objective function (e.g., by reducing error or loss) when it is used for a split. Features with higher gain values are considered more important, since they contribute more to improving the model's accuracy.
%
%
Another metric is {\em cover}, which reflects how many training samples are influenced by a split using a particular feature. In practice, this means that features used higher up in the trees, which affect larger groups of samples, tend to have higher cover values.
The last metric is {\em frequency}, which simply counts how often a feature is used to split the data across all trees in the model. A higher frequency indicates that the feature is frequently chosen as a useful splitting criterion.

Although these feature importance metrics provide some insight into which features the model relies on most, they miss several key aspects. First, none of these metrics explains {\em how} a feature influences the model's prediction. In other words, they do not reveal how a change in the value of a specific feature would affect the output. For example, would an increase in a particular feature make a client more likely to default, or would it make the loan appear safer? 
Secondly, these metrics do not offer intuition about the model's behavior in the absence of a feature. Would the predictions of the model change significantly if certain features were excluded? 
Lastly, they fail to capture the impact of interactions between features on the final prediction. 

These limitations underscore the need for more advanced tools such as SHAP to interpret the inner workings of complex models. 
As noted earlier, especially the Tree SHAP method is particularly designed for fast calculation of Shapley values for tree-based models.

XAI models  that combine ensemble tree-based algorithms with SHAP have become increasingly prevalent in recent credit scoring literature, as demonstrated by \textcite{misheva_explainable_2021}, \textcite{zhu_ensemble_2024}, and \textcite{alonso_understanding_2021}.

\subsection{Glass-Box Models from Recent Years}

At the other end of the transparency spectrum are the glass-box models. This category includes traditional methods such as regression and decision trees, as well as more recent models that incorporate additional complexity while maintaining transparency. 

Our literature review identifies two glass-box models from recent years that show promising performance: Explainable Boosting Machine (EBM) and 
Explainable Boosting Machine (EBM) and Penalized Logistic Tree Regression (PLTR). EBM, developed by \textcite{nori_interpretml_2019}, is a supervised machine learning algorithm specifically designed to balance interpretability and predictive power. It offers model intelligibility while achieving accuracy comparable to state-of-the-art algorithms such as XGBoost. PLTR, introduced by \textcite{dumitrescu2022machine}, uses predictors derived from shallow decision trees as inputs to a penalized logistic regression, offering a hybrid approach that enhances both interpretability and predictive performance.

\subsubsection{Explainable Boosting Machine (EBM)}

The EBM algorithm belongs to the family of generalized additive models (GAMs), which extend generalized linear models (GLMs) by modeling the expected value of the target variable as a sum of smooth, non-linear functions of the predictors. 

EBM trains the model iteratively in a round-robin fashion: in each iteration, a small tree is fitted to capture the effect of one feature on the residuals, and the residuals are updated accordingly. This process is repeated across the features one by one, completing one round-robin pass through the dataset. A small learning rate ensures that the order in which features are included does not significantly influence the final model. As a result, the training process requires many iterations, as many as ten thousand, to reach convergence. The result is an ensemble of small trees per feature, which can be aggregated collectively to capture the relationship between the feature and each target variable.

In mathematical terms, the prediction function reads
\begin{equation}\label{eq:EBM}
g(E[Y]) = \beta_0 + \sum_{j=1}^{d} f_j(x_j),\end{equation} where $g$ is a link function and each $f_j$ denotes a smooth function representing the effect of $j$-th feature, i.e. $x_j$, on the target variable. 
%
Worth noting that, the difference between Eq. (\ref{eq:EBM}) and Eq. (\ref{eq:XGBoost}) is that the functions involved in EBM prediction are uni-variate functions, while those involved in XGBoost take multiple features as inputs. Each marginal function $f_j(\cdot)$ is stored after training and can be visualized in plots. These stored functions are eventually used for predictions later on. 

Building on this concept, \textcite{lou_accurate_2013} introduced Generalized Additive Model plus Interactions (GA$^2$M), which incorporates pairwise feature interactions as follows: \begin{equation}\label{eq:GA2M}g(E[Y]) = \beta_0 + \sum_{j=1}^d f_j(x_j) + \sum_{1 \le j < q \le d} f_{j,q}(x_j, x_q)
\end{equation}
Despite the added complexity, GA$^2$Ms remain interpretable due to their additive structure and the use of small CARTs or shallow trees to model each term. 
EBM can identify and incorporate feature interactions by training pairwise interaction terms, if enabled, using a similar iterative process but built on Eq. (\ref{eq:GA2M}). 

Finally, the EBM library generates visualizations that illustrate feature importance and interactions that provide insights into the model's inner workings.

Testing results in \cite{dessain_ebm_2023} show that EBM performs comparably to  black-box gradient boosting models, such as XGBoost. This result appears to be consistent in models used outside the scope of credit scoring and finance \parencite{caruana_intelligible_2015}.

\subsubsection{Penalized Logistic Tree Regression (PLTR)}

The other powerful and explainable model is called Penalized Logistic Tree Regression, or PLTR for short. It is introduced in \cite{dumitrescu2022machine} for a credit scoring model. 
The model is constructed in two steps. 

In the first step,  binary features are created using short-depth decision trees with one or two splits. Based on single-split trees, each feature $x_j$ generates a new binary feature $\nu^{(j)}$ for logistic regression, which takes the value 1 if the $j$-th feature of the $i$-th sample exceeds the threshold from the single-split tree, and 0 otherwise. From two-split trees involving two features $x_j$ and $x_q$, with $x_j$ assumed to be more informative, we first define $\nu^{(j)}$ as in the single-split case, then we define a second binary feature $\xi^{(j,q)}$, which takes the value $1$ if $x_j$ is below its threshold and $x_q$ is above its threshold. The final set of features used in the model thus consists of original set of features ${x_j}$, binary features from the single-split  trees $\nu^{(j)}$, and binary features from the two-split trees $\xi^{(j,q)}$.

In the second step,  a penalized logistic regression is applied to this expanded set of features. 
Recall the logistic regression log-likelihood in Eq. (\ref{eq:log-loss}), we have
\begin{equation}
\mathcal{L}(\mathbf{\theta}) = \frac{1}{n} \sum_{i=1}^n \Bigl[  y_i \ln \left[\sigma\left(z(\mathbf{\theta},\mathbf{\tilde{x}}_i)\right)\right] 
+  \left(1-y_i\right) \ln \left[1-\sigma\left(z(\mathbf{\theta},\mathbf{\tilde{x}}_i)\right)\right] \Bigr],
\end{equation}
where $\mathbf{\tilde{x}}_i$ is the extended feature vector from the $i$-th sample, containing all $x_j$, $\nu_i^{(j)}$ and $\xi_i^{(j,q)}$ for all  $j \neq q$, and 
$\mathbf{\theta}$ 
is the respective set of model parameters.

To prevent over-fitting, we introduce regularization or penalization by adding a penalty term to the negative log-likelihood function:
$$
\tilde{\mathcal{L}}(\mathbf{\theta})=-\mathcal{L}(\mathbf{\theta})+\lambda P(\mathbf{\theta}),
$$
where $P(\mathbf{\theta})$ is the additional penalty term and $\lambda$ is the tuning parameter controlling the strength of the penalty. There are many choices of $P(\mathbf{\theta})$,  \cite{zou2006adaptive} proposed  the adaptive lasso estimator.


\section{Comparing the Models, Is There a Silver Bullet?}

As summarized in the discussion paper from \cite{eba2023machine}, a central challenge in applying machine learning techniques to regulated areas such as IRB PD modeling lies in their complexity, which hinders explainability.
In particular, that discussion paper recommends institutions to avoid, among others,	including an excessive number of explanatory drivers or drivers with no significant predictive information, and overly complex modeling choices if simpler approaches yielding similar results are available.

While existing XAI approaches, such as combining black-box models with post-hoc interpretability tools like SHAP, provide some insights and certain level of understanding of the model behaviors, they do not automatically meet those guidelines. These models typically involve a large number of features. Nonlinear behaviors combined with numerous feature interactions are difficult to interpret and communicate, particularly for model owners and regulatory supervisors. Even when using inherently interpretable models like EBM, introducing feature interactions can result in patterns that defy financial intuition. 

To address this gap, we propose a hybrid approach that leverages the feature rankings produced by post-hoc interpretability tools applied to black-box models, and then uses these insights to construct interpretable, glass-box models. This method maintains predictive performance while significantly reducing model complexity and feature count. The resulting model based on the proposed approach is potentially more suitable for regulated environments than existing XAI models.

In the following subsections, we present the proposed method, describe the dataset and experimental setup, compare our lightweight models against benchmarks in terms of both accuracy and interpretability, and finally show how domain expertise and further analyses (e.g., feature interactions and correlations) can be used to fine-tune the model.

\subsection{Our Proposal}\label{sec:method}

We propose the following approach:
\begin{itemize}
    \item[-] Step 1: train a well performing black-box machine learning model using the full feature set. This we name as the ``base'' model, which serves both as a performance benchmark as well as the basis for feature selection for step 2. 
    \item[-] Step 2: rank features by their importance in the base model, by for example applying a post-hoc interpretation tool like  SHAP.
    \item[-] Step 3: select a few, such as 10 or 20, top-ranked features to train a glass-box model, which can be EBM or PLTR for example.
    \item[-] Step 4: based on interaction analysis, correlation analysis, and/or expert opinion, refine the selection of top-ranked features and repeat step 3.
\end{itemize}

The result is explainable and technically transparent models that better align with the expectations of both regulators and practitioners in the credit risk domain.

To evaluate the potential of this approach for credit scoring, we used the top-performing classifiers introduced in Section~\ref{sec:XGBoost} as benchmarks. 

Note that, as our goal is to propose a general method to simplify and "glassify" black-box models while preserving their predictive power, we did not extensively optimize the base models. Nor did we aim to reconfirm that tree boosting algorithms outperform logistic regression in credit scoring - an already well-established result.

\subsection{Data Preparation and Base Models}

Our experiments were based on data from the Lending Club peer-to-peer lending platform, which operated until the end of 2020. This dataset has been widely used in prior studies on explainable machine learning methods, allowing us to compare against established methods and focus on persistent challenges in interpreting ML models in financial applications.

We preprocessed Lending Club data following similar steps to those taken in \textcite{ariza-garzon_explainability_2020}. As for  the target variable, we decided to consider only `Fully Paid' and `Charged Off' categories of the variable \textit{loan\_amount}, encoding them as 0 and 1 respectively. We also took the average of \textit{fico\_range\_high} and \textit{fico\_range\_low} to create the feature \textit{fico\_range\_low}, while removing the original features. The features \textit{home\_ownership}, \textit{purpose}, \textit{addr\_state} and \textit{emp\_length} were encoded using one-hot encoding. Where necessary, missing values were imputed with the mean  of the corresponding feature.

The split between training data and test data was done based on \textit{issue\_d} feature, such that the train data consisted of data up to July 2015, and the test set consisted of data from August 2015 to December 2018. The overall default rate observed in the data was around 20\%, and the total number of features was 87.
Using the preprocessed data, we built three baseline models employing the algorithms previously discussed: logistic regression (LR), XGBoost (XGB) and EBM. Since the data at hand is imbalanced - in the sense that the defaulting class is significantly smaller than the non-defaulting class, we assigned class weights to each  model. This approach typically enhances the model's ability to correctly classify the minority class, often at the expense of performance on the majority class. However, in the context of credit risk modeling, prioritizing the accurate identification of default events is both acceptable and desirable.

For evaluating the performance of the different models, we used multiple performance metrics. However, we placed greater emphasis on metrics that appropriately account for the performance on the minority class. In addition to balanced accuracy, we included AUROC (Area Under the Receiver Operating Characteristic curve), F1 scores and AUPRC (Area Under the Precision Recall Curve). 


Table~\ref{tab:Base_models_per} summarizes the performance of the models which we aim to preserve in our simplified, interpretable approach.

\begin{table*}
    \centering
    \begin{tabular}{ccccc}
         Model&  AUPRC&  AUROC&  F1 score& Balanced Accuracy\\
         \hline
         LR&  0.3389&  0.6653&  0.42& 0.6187\\
         XGB&  0.3436&  0.6687&  0.4160& 0.6203\\
 EBM& 0.3518& 0.6744& 0.4211&0.6251\\
    \end{tabular}
    \caption{Performance of base models (with all features included)}
    \label{tab:Base_models_per}
\end{table*}

\subsection{Issue in Interpreting Base Models}


Given the volume of features and the complex, implicit interactions captured by the model, validating whether the model's decisions align with domain knowledge becomes a daunting task. Top features identified by SHAP can help in explaining the scoring result for each individual client. However, it is challenging to explain how much the interactions between features affect the final score and why a large number of features are still needed.


%
%
%
%

\begin{figure}[ht]
    \centering
    \includegraphics[width=0.8\linewidth]{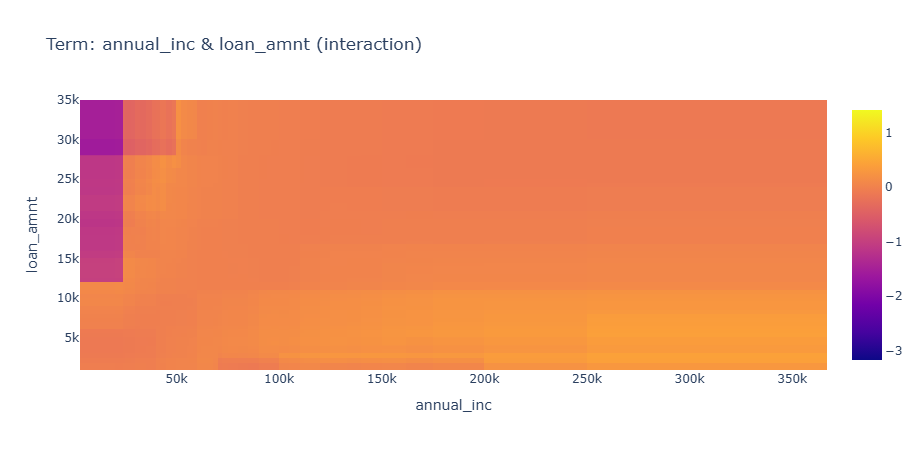}
    \caption{Loan amount and annual income feature interaction (from SHAP)}
    \label{fig:ebm_feature_interaction}
\end{figure}

Moreover, even when using inherently interpretable models like EBM, introducing feature interactions can result in patterns that defy financial intuition. As shown in Figure~\ref{fig:ebm_feature_interaction}, the model suggests that applicants with lower loan amounts and higher salaries are more likely to default than those with higher loan amounts and lower salaries. This contradicts basic financial reasoning. However, when we examine the individual effects of these two features, \textit{loan amount} and \textit{annual income}, separately, their impact aligns much more closely with human expectations and established financial logic.

\begin{figure}[ht]
    \centering
    \includegraphics[width=1\linewidth]{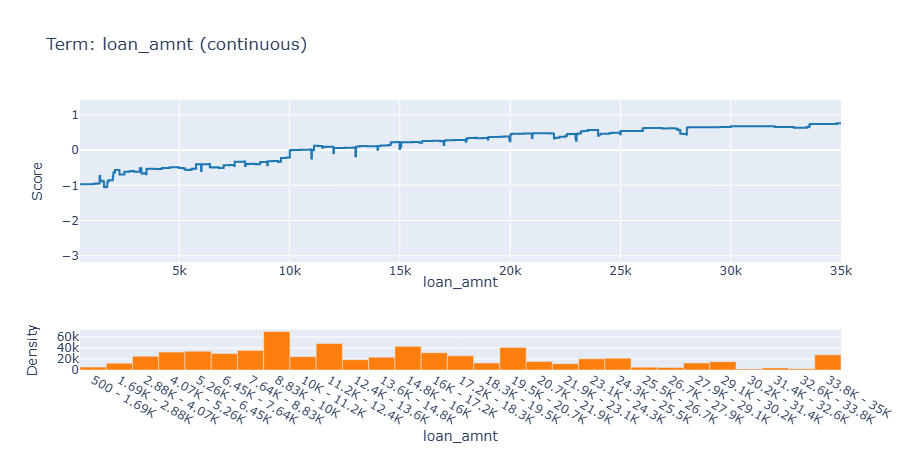}
    \caption{Loan amount attribution (from SHAP)}
    \label{fig:loan_amnt_attr}
\end{figure}

\begin{figure}[ht]
    \centering
    \includegraphics[width=1\linewidth]{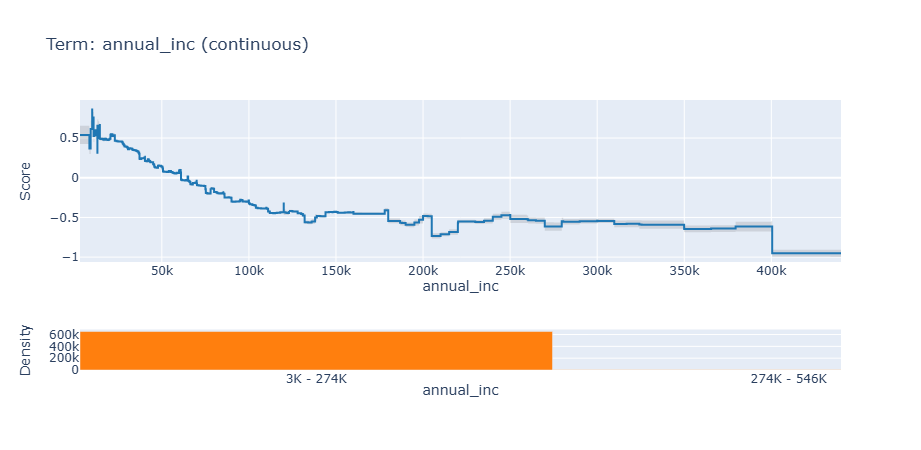}
    \caption{Annual income attribution (from SHAP)}
    \label{fig:annual_inc_attr}
\end{figure}

As illustrated in Figure~\ref{fig:loan_amnt_attr} and Figure~\ref{fig:annual_inc_attr}, we observe a clear and intuitive trend: higher loan amounts are generally associated with a higher probability of default, while higher annual incomes correspond to a lower probability of default. This motivates us to largely reduce the number of features, aligning with the guidelines given by earlier mentioned discussion paper, so as to minimize feature interactions. 

\subsection{Feature Importance}


Having built the three base models, we now proceed to Step 2 as described in Section~\ref{sec:method}, where we rank the features based on their importance.

For the logistic regression (LR) model, feature importance is determined using the model coefficients after normalizing the input features. In the case of XGBoost, we rely on tree-SHAP values to obtain feature attributions. For the EBM model, we use its inherent feature importance scores. The resulting feature rankings are summarized in Table~\ref{tab:base_models_top_features}. They are largely consistent across all three models, with only minor differences in the ordering (note that for EBM, interaction terms were initially excluded from the ranking). 

\textbf{Remark}: Among the three, XGBoost provided the fastest computation of feature importance, making it a practical choice for feature selection.

\begin{table}
    \centering
    \begin{tabularx}{\textwidth}{cXXX}
         Rank & LR & XGB & EBM \\
         \hline
         1 & fico\_range\_high& fico\_range\_high& loan\_amnt \\
         2 & loan\_amnt & loan\_amnt & fico\_range\_high\\
         3 & dti & annual\_inc & annual\_inc\\
         4 & annual\_inc & dti & purpose\_credit\_card\\
         5 & home\_ownership\_RENT & purpose\_credit\_card & dti\\
         6 & purpose\_credit\_card & home\_ownership\_RENT & purpose\_debt\_consolid. \\
         7 & emp\_length\_nan & home\_ownership\_MORT. & home\_ownership\_RENT \\
         8 & purpose\_debt\_consolid. & purpose\_debt\_consolid. & home\_ownership\_MORT. \\
         9 & purpose\_other & emp\_length\_nan & emp\_length\_nan\\
         10 & adde\_state\_NY & adde\_state\_NY & adde\_state\_NY\\
         \hline
    \end{tabularx}
    \caption{Top features in different models}
    \label{tab:base_models_top_features}
\end{table}

\subsection{Light-weight Glass-box Models}

After obtaining the feature importance rankings discussed above, we proceeded to build glass-box models using subsets of the top-ranked features of varying sizes, i.e. Step 3 of our approach as described in Section \ref{sec:method}. The glass-box models we considered were EBM and PLTR.

\begin{figure}[ht]
    \centering
    \includegraphics[width=1\linewidth]{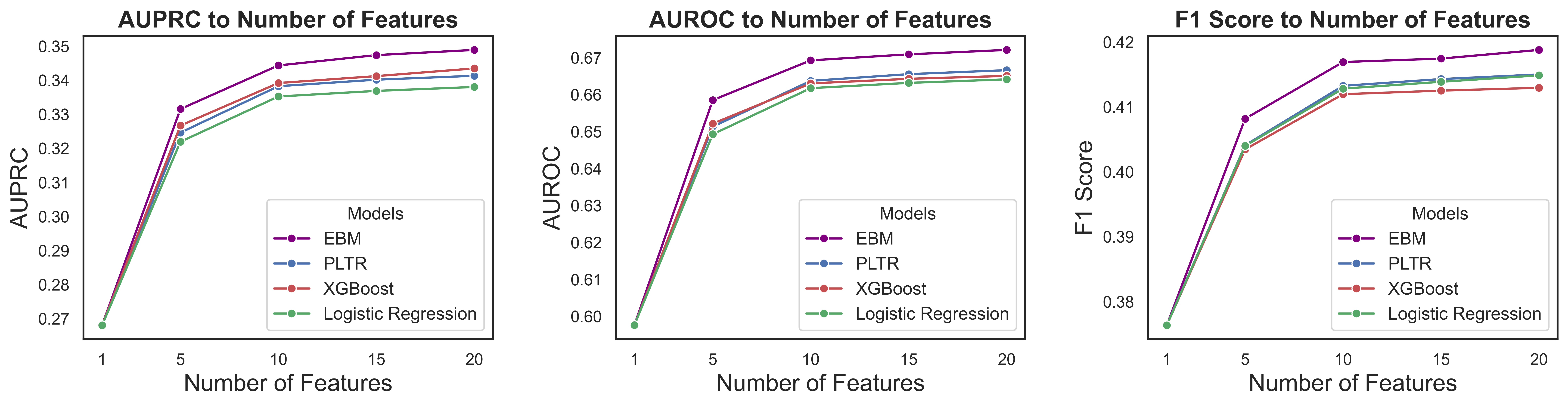}
    \caption{Sensitivity analysis: Performance w.r.t. the number of (top) features included}
    \label{fig:reduced_models_plot}
\end{figure}

By plotting performance metrics against the number of included features, we could identify a practical cut-off point for constructing a reduced model. Figure~\ref{fig:reduced_models_plot} displays the progression of AUPRC, AUROC, and F1 scores as a function of the number of features. In the case of the Lending Club data, we observe that adding more than 10 features yields no substantial improvement in model performance. This insight allowed us to reduce the dimensionality from 86 features to just 10, without significantly compromising predictive power.

Notably, EBM consistently achieves performance that is at least on par with XGBoost, while offering significantly greater transparency. PLTR also improves interpretability relative to XGBoost but underperforms compared to both EBM and XGBoost in terms of predictive accuracy. As a result, EBM emerges as the most favorable model in our study, providing a strong balance between explainability and performance.

\subsection{Refinements}
Once the reduced EBM model is obtained, both global and local analyses can be conducted with ease. Since EBM stores all marginal functions used in its predictions and the reduced EBM model uses a limited number of features, interpreting the model becomes straightforward. These insights can, in turn, be used to further refine and enhance the model.


\subsubsection{Feature Interaction Analysis}

As shown earlier, even pairwise interactions can lead to counterintuitive or misleading conclusions, like in the case of the Lending Club data. 

\begin{figure}[ht]
    \centering
    \includegraphics[width=1\linewidth]{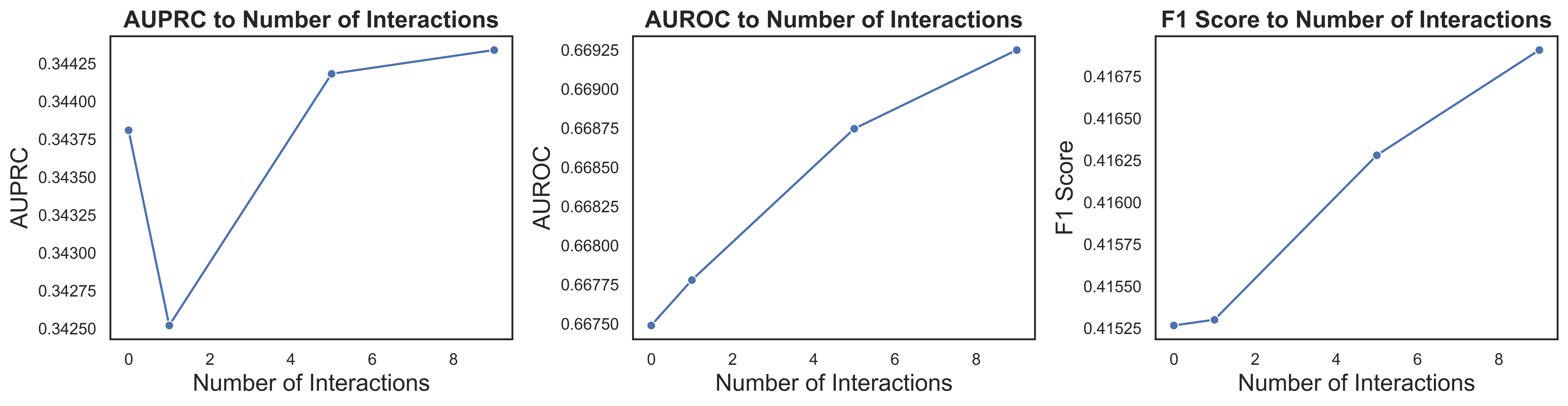}
    \caption{Sensitivity analysis: EBM performance w.r.t the number of pairwise interactions}
    \label{fig:ebm_interactions}
\end{figure}

To further assess the impact of pairwise interactions on model performance, we built a series of ancillary models, each incorporating a different number of pairwise interactions, ranging from zero up to nine.
Figure~\ref{fig:ebm_interactions} summarizes the different performance metrics as a function of the number of pairwise interactions from these ancillary models.  The maximum improvement was measured using F1 score and amounted to 0.4\% when all nice pair-wise interactions were included, which is a very marginal gain. This analysis suggests that, in some cases, practitioners may confidently exclude all feature interactions and still obtain a glass-box model that outperforms the logistic regression benchmark.

\subsubsection{Correlation Analysis}

It is important to note that while SHAP analysis provides a robust, data-driven estimation of feature importance, it also has certain limitations. A notable weakness lies in its handling of correlated features. When features are highly correlated, SHAP sometimes distributes a positive SHAP value to one of the correlated pair, while a negative SHAP value to the other, or distribute the importance across them. This makes it difficult to fully trust the top feature list selected using SHAP. 
To mitigate this limitation, we recommend to combine SHAP with correlation analysis.

We computed the correlation matrix among the top 25 most important features identified by SHAP. For each pair of highly correlated features, we removed the lower-ranked one if it was not among the top 10 most important features. This process was repeated until we obtained a revised set of 20 features. It is important to note that we never removed any of the original top 10 SHAP-ranked features, even if they were part of a highly correlated pair. Empirical results show that removing one of the top 10 features often leads to a decline in model performance, supporting the decision to keep them. The results in the Table \ref{tab:xgb_performance_correlation} also suggested that this method can imporve the performance of the model. 

\begin{table*}[ht]
    \centering
    \begin{tabular}{cccccc}
         Model & Method &  AUPRC&  AUROC&  F1 score& Balanced Accuracy\\
         \hline
        XGB & Original&  0.3412&  0.6641&  0.4139& 0.6185\\
         XGB & Correlation Analysis&  0.3418&  0.6658&  0.4145& 0.6190\\
         EBM & Original&  0.3487&  0.6720&  0.4186& 0.6226\\
         EBM & Correlation Analysis&  0.3493&  0.6725&  0.4193& 0.6233\\

 \hline
    \end{tabular}
    \caption{Performance comparison for XGBoost and EBM: using original top 20 features without correlation analysis vs using top 20 features after correlation analysis}
    \label{tab:xgb_performance_correlation}
\end{table*}




\section{Conclusion}

In this study, we proposed an approach that has the potential to assist banks and financial institutions in developing more transparent and reliable models with fewer features, meeting both practical needs and regulatory expectations for Probability of Default (PD) models.

Using the Lending Club dataset, we demonstrated that, following our approach, it is possible to construct a lightweight glass-box model comprising only a small subset of features, while achieving predictive performance comparable to a black-box model utilizing the full feature set.

For future research, we recommend testing the effectiveness and robustness of this approach across diverse datasets to validate its generalizability and applicability in varied financial contexts.


\section*{Declaration of Interest}
The authors report no conflicts of interest. The authors alone are responsible for the content and writing of the paper.



\printbibliography

\end{document}